\documentclass[10pt,twocolumn,letterpaper]{article}
\PassOptionsToPackage{numbers, compress}{natbib}
\usepackage{cvpr}
\usepackage[utf8]{inputenc} 
\usepackage[T1]{fontenc}    
\usepackage{url}            
\usepackage{booktabs}       
\usepackage{amsfonts}       
\usepackage{nicefrac}       
\usepackage{microtype}      
\usepackage{times}
\usepackage{epsfig}
\usepackage{graphicx}
\usepackage{amsmath}
\usepackage{amssymb}
\usepackage{stmaryrd}
\usepackage[export]{adjustbox}
\usepackage{balance}
\usepackage{multirow}
\usepackage{float}
\usepackage{wrapfig}
\usepackage{amsthm}
\usepackage{algorithm,algorithmicx,algpseudocode}
\usepackage{bm,xspace}
\usepackage{comment}
\usepackage{verbatim}
\usepackage{etoolbox,siunitx}
\usepackage{calc}
\usepackage{pifont,hologo}
\usepackage[usenames, dvipsnames]{color}

\usepackage{hyperref}
\hypersetup{pagebackref=true,breaklinks=true,letterpaper=true,colorlinks,bookmarks=false}

\cvprfinalcopy

\newcommand{\ra}[1]{\renewcommand{\arraystretch}{#1}}

\ifcvprfinal\fi

\begin{document}

\title{Future Video Synthesis with Object Motion Prediction}

\author{Yue Wu\\
HKUST
\and
Rongrong Gao\\
HKUST
\and
Jaesik Park\\
POSTECH
\and
Qifeng Chen\\
HKUST
}

\maketitle

\begin{abstract}
We present an approach to predict future video frames given a sequence of continuous video frames in the past. Instead of synthesizing images directly, our approach is designed to understand the complex scene dynamics by decoupling the background scene and moving objects. The appearance of the scene components in the future is predicted by non-rigid deformation of the background and affine transformation of moving objects. The anticipated appearances are combined to create a  reasonable video in the future. With this procedure, our method exhibits much less tearing or distortion artifact compared to other approaches. Experimental results on the Cityscapes and KITTI datasets show that our model outperforms the state-of-the-art in terms of visual quality and accuracy.
\end{abstract}

\section{Introduction}
Can an artificial intelligence system predict a photorealistic video conditioned on past visual observation? With an accurate video prediction model, an intelligent agent can plan its motion according to the predicted video. Future video generation techniques can also be used to synthesize a long video by repeatedly extending the future of the video. Video prediction has been adopted in various applications such as sensorimotor control, autonomous driving, and video analysis~\cite{Finn2016,Srivastava2015,NextSegmPredICCV17,highfideity}.  

Video prediction has not been solved yet, especially if we need to synthesize frames of an extended period. Existing methods tend to generate blurry and distorted images where rigid objects are usually bent and spread. This issue indicates that it is necessary to consider several aspects: forecasting the motion of dynamic objects, creating new visual data for unveiled regions, finding spatio-temporal relationships when two objects overlap, and so on. Therefore, to generate realistic future video, understanding essential information such as semantics, shape, or dynamics of the scene is necessary. 

Most existing methods tackle the video prediction task by generating future video frames one by one in an unsupervised fashion~\cite{prednet,xue2016visual,Villegas2017,DentonFergus2018}. These approaches synthesize future frames at the pixel level without explicit modeling of the motions or semantics of the scene. Thus, it is difficult for the model to grasp the concept of object boundaries to create different movements for different objects. 
For instance, a moving car should be treated individually instead of modeling a car and background scene as a whole. Recently, Wang et al.~\cite{Wang2018} propose a general video-to-video translation model (vid2vid) that demonstrates future video prediction as a sub-task. The model takes semantic maps in the past and estimates future semantic maps to synthesize the next video frame. With this idea, the generated video can preserve the structure of objects better, but the shape of objects deforms unnaturally in the long term. To synthesize more realistic future videos, we find that the explicit modeling of object trajectories is highly beneficial.

The key idea of our video prediction model is that we synthesize future video frames conditioned on predicted object trajectories. The trajectory of an object is defined as its 2D pixel location in each video frame. In particular, we identify each dynamic object and predict its moving path, scale change, and shape in the future. Object appearance in the next few frames can be roughly approximated by applying an affine transformation on the object segment in the last input frame. In this way, appearance is highly regularized and avoids unexpected deformation. For the background with static objects, we directly predict a motion field between the last frame and each future frame. Then we warp the background image with the estimated motion field. In this background image in the future, dynamic objects are located. Since the future background images may contain missing regions due to occlusion, we apply refinement steps to complete missing areas and harmonize components. Our experiments indicate that our approach can synthesize future videos that are more photo-realistic than state-of-the-art video prediction methods.

\begin{figure*}[t!]
\hspace{-4mm}
\setlength\tabcolsep{2pt}
\begin{tabular}{ccccc}
\rotatebox{90}{\hspace{7mm}$t+1$} & \includegraphics[width=0.24\linewidth]{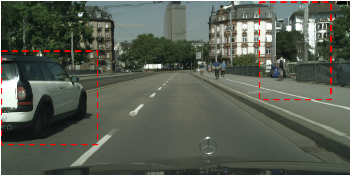}  & \includegraphics[width=0.24\linewidth]{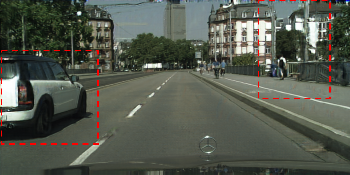}  & \includegraphics[width=0.24\linewidth]{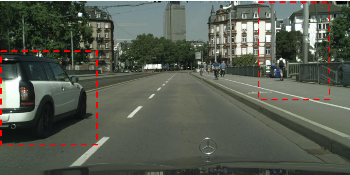} &
\includegraphics[width=0.24\linewidth]{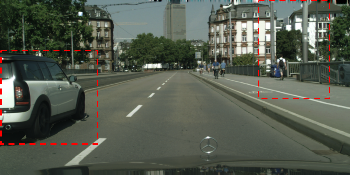}\\
\rotatebox{90}{\hspace{7mm}$t+5$} & \includegraphics[width=0.24\linewidth]{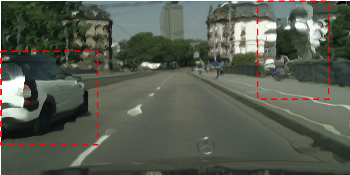}  & \includegraphics[width=0.24\linewidth]{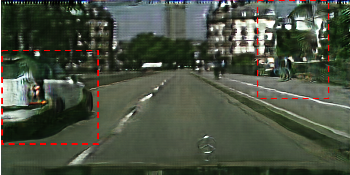}  & \includegraphics[width=0.24\linewidth]{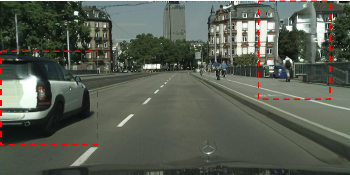} &
\includegraphics[width=0.24\linewidth]{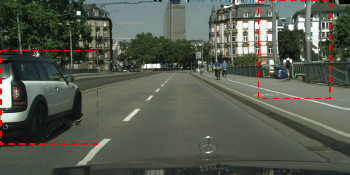}\\
\rotatebox{90}{\hspace{7mm}$t+10$} & \includegraphics[width=0.24\linewidth]{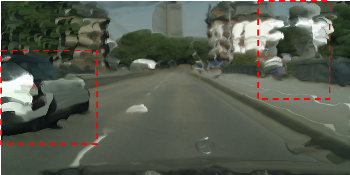}  & \includegraphics[width=0.24\linewidth]{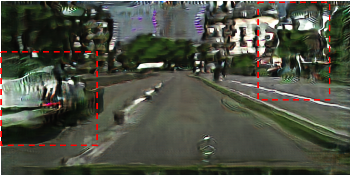}  & \includegraphics[width=0.24\linewidth]{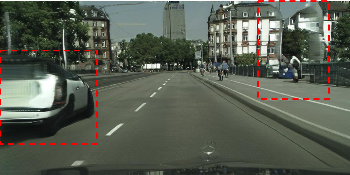} &
\includegraphics[width=0.24\linewidth]{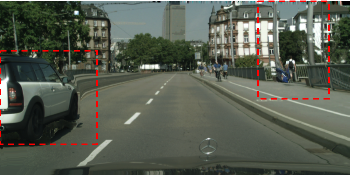}\\
\rotatebox{90}{\hspace{7mm}$t+1$} & \includegraphics[width=0.24\linewidth]{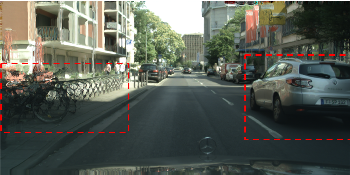}  & \includegraphics[width=0.24\linewidth]{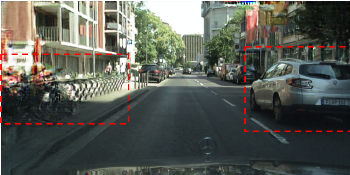}  & \includegraphics[width=0.24\linewidth]{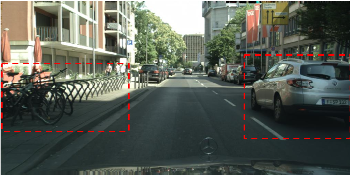} &
\includegraphics[width=0.24\linewidth]{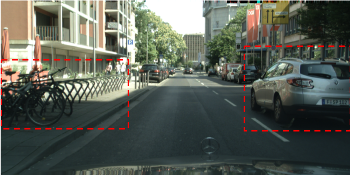}\\
\rotatebox{90}{\hspace{7mm}$t+5$} & \includegraphics[width=0.24\linewidth]{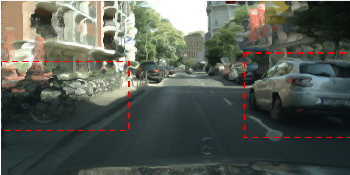}  & \includegraphics[width=0.24\linewidth]{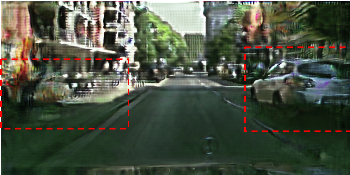}  & \includegraphics[width=0.24\linewidth]{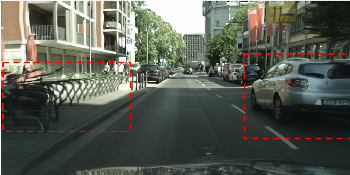} &
\includegraphics[width=0.24\linewidth]{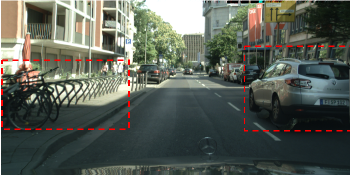}\\
\rotatebox{90}{\hspace{7mm}$t+10$} & \includegraphics[width=0.24\linewidth]{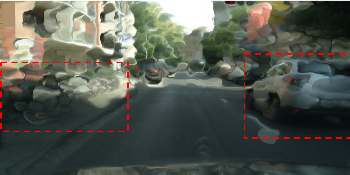}  & \includegraphics[width=0.24\linewidth]{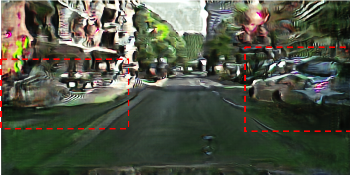}  & \includegraphics[width=0.24\linewidth]{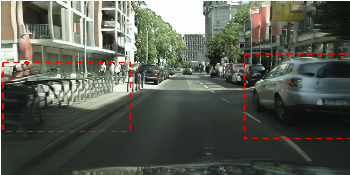} &
\includegraphics[width=0.24\linewidth]{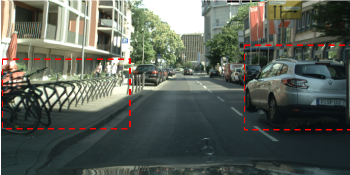}\\
& Voxel-Flow~\cite{liu2017voxelflow} & MCNet~\cite{villegas17mcnet} & vid2vid~\cite{Wang2018} & Ours \\
\end{tabular}
\caption{Results of predicting the frames $t+1$, $t+5$ , and $t+10$ on the Cityscapes dataset~\cite{Cordts2016Cityscapes}.}\label{fig:cityscapes_compare}
\end{figure*}

\section{Related Work}
Future frame synthesis is initially studied at the patch level~\cite{Sutskever2009}. Recent advances in the future prediction from image sequence can be classified into the three-fold.

\vspace{2mm}
\noindent\textbf{Single image prediction.}
This class of works synthesizes a single frame for the next time step. Patraucean~\etal~\cite{Patraucean2016} use a convolutional version of long short-term memory. Lotter~\etal~\cite{prednet} introduce a predictive coding network, and Byeon~\etal~\cite{Byeon2018} improve image quality using parallel multi-dimensional long short-term memory (LSTM). Liang~\etal~\cite{Liang2017} design the generative adversarial loss~\cite{Goodfellow2014} for both on predicted optical flow and synthesized image to enforce consistency explicitly. Liu~\etal~\cite{Liu2018} introduce an efficient atomic operator to predict the next frame in an unsupervised manner. 

\vspace{2mm}
\noindent\textbf{Long-term prediction.}
More recently, long-term video prediction becomes an active research area. Srivastava~\etal~\cite{Srivastava2015} use LSTM to encode and decode video sequence. Denton and Fergus~\etal~\cite{DentonFergus2018} introduce an approach that produces plausible frame predictions with stochastic latent variables to generate sharp frames. Mathieu~\etal~\cite{Mathieu2015} propose a multi-scale approach by reducing blur artifact with the aid of mean squared error loss.  Lee~\etal~\cite{Lee2018} combine the latent variable for stochastic reasoning and adversarial loss for photo-realistic image synthesis. Wichers~\etal~\cite{Wichers2018} introduce a hierarchical approach without using ground truth annotation of high-level structures. A probabilistic approach by Xue~\etal~\cite{xue2016visual} synthesizes various motions from a single image. Villegas~\etal~\cite{villegas17mcnet} and Reda~\etal~\cite{Reda2018} involve motion encoder to explicitly regard foreground motion. Wang~\etal~\cite{Wang2018} propose an advanced framework that can synthesize long-term video. The power of this approach comes from a concrete design of generative adversarial loss for image domain and temporal domain. Ye~\etal~\cite{ye2019Compositional} proposed a pixel-level future prediction approach given a single image with the prediction of future states of independent entities. 
Liu~\etal~\cite{Liu2019VRNN} used variational recurrent neural networks with higher capacity likelihood models.
Hang~\etal~\cite{Hang2019Dis} introduced a confidence-aware warping operator to predict occluded area and disoccluded area separately.
Ho~\etal~\cite{Ho2019SMENET} proposed a parametric video prediction approach based on a sparse motion field.

\vspace{2mm}
\noindent\textbf{Other tasks.}
In addition to the next image or long-term image synthesis, additional tasks, including scene semantics or motion of dynamic objects, have been studied. A method by Walker~\etal~\cite{Walker2016} predicts the movement of the foreground object from a single image, and recent works~\cite{Alahi2016, Bhattacharyya2018, Gupta2018, kitani2012activity, Ma2019, Yagi2018, Djuric2018, Sadeghian2018} demonstrate that human movements or trajectories can be estimated successfully from the real dataset. Vondrick~\etal~\cite{Vondrick2016} synthesize a one-second video from weakly annotated natural videos using a network understanding dynamics of foreground object and motion classes. Jin~\etal~\cite{jin2017predicting} propose a new fully convolutional network for predicting semantic label and optical flow for a next frame. 

Our approach is in line with recent works~\cite{denton2017unsupervised, villegas17mcnet, luo2017unsupervised, Qi2019} that decouple stationary and moving part of the scene. We incorporate high-level semantics and instances to consider movements of individual foreground objects explicitly. As shown in the paper, the synthesized frames are more realistic than previous state-of-the-art on complex scenes, such as real-world driving videos.

\section{Model}
\noindent\textbf{Problem definition.}
Let $x_{i}$ be the video frame at time step $i$, $s_{i}$ and $e_{i}$ be the corresponding semantic and instance map of $x_{i}$, and $f_{i}$ be the motion field (or optical flow) from frame $x_{i}$ to frame $x_{i+1}$. Then our video prediction task could be formulated as follows. Given input video frames $x_{i}$, semantic maps $s_{i}$, instance maps $e_{i}$ from time-steps $i=\{1,\cdots,t\}$ and optical flows between consecutive frames, predict the future video frames $x_{i}$ for $i=\{t+1,\cdots,T\}$. $t$ is the index to the last input frames, and $T$ is the index of the last prediction frame. To solve this problem, we propose a separate-predict-composite approach to produce realistic future frames. 

\begin{figure*}[t!]
\hspace{-4mm}
\centering
\includegraphics[width=1\linewidth]{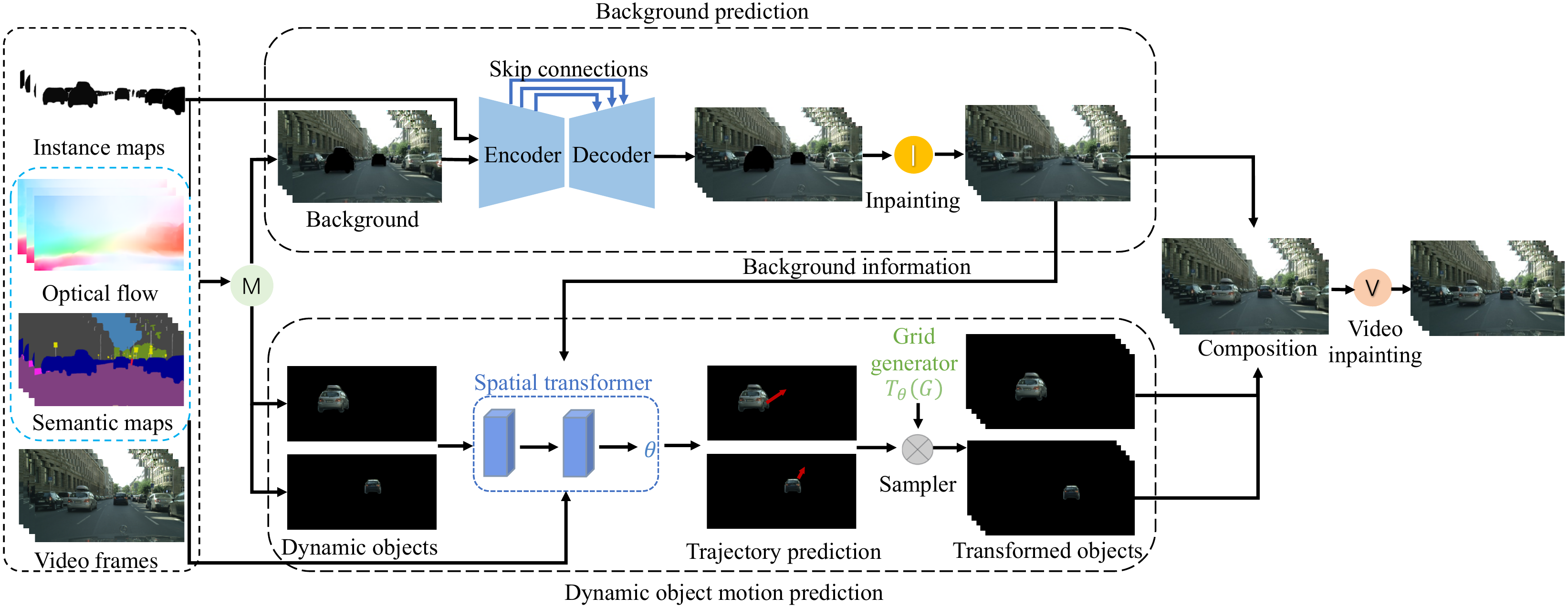}
\vspace{1mm}
\caption{
Overview of the proposed architecture. We use a dynamic object detection model $M$ to separate moving objects and static background. The missing foreground area in the generated future background is inpainted using the inpainting model $I$. By providing the background images for the future, we apply a spatial transformer to predict moving objects. After that, we composite the foreground and background images and use a video inpainting module $V$ to inpaint occluded area.
}
\label{fig:overview}
\end{figure*}

\vspace{2mm}
\noindent\textbf{Overview.}
To begin with, we attempt to classify objects into dynamic and static ones to trace and handle various motions in the scene effectively. 
We train a moving object detection network to classify moving objects and static scenes. 
This idea is different from previous approaches that divide frames to the foreground and background region based on semantic class.
After obtaining dynamic and static regions, we predict the optical flow of the static scene and warp the last input video frame to get future frames of the static scene. 
Then, we use a background-aware spatial transformation network (STN) to predict the motion of dynamic objects. The holes in warped static scenes are filled using an image inpainting method~\cite{yu2018generative}. The warped images serve as the future background information for the STN network. 
The estimated static and dynamic scene is composed in the last stage to generate a seamless image. Fig.~\ref{fig:overview} illustrates the proposed pipeline.

\subsection{Moving object detection}
There are two major causes for the appearance change between continuous frames. The first one is the dynamic motion of moving objects, and the second one is the ego-motion of the camera. To handle such a scene effectively, we train a moving object detection network to identify moving objects and static scenes. Based on Cityscapes-Motion dataset and KITTI-Motion dataset~\cite{Valada_2017_IROS} that provide annotations of moving areas, we build an encoder-decoder architecture to detect moving regions, with ResNet50 as the backbone~\cite{resnet}. The input of the network is observed sequences of frames, semantic maps, instance maps, and optical flow between consecutive frames. The output of the network is a binary mask to indicate the region of the moving object.

\subsection{Background prediction}
With the identified moving objects, the pipeline handles the static motion of the scene that is predicted by an optical flow network. The network predicts the forward and backward optical flow between the last observed frame $x_{t}$ and each future frame\footnote{Backward optical flow is used for warping $x_t$ to the future because this can avoid warping artifacts.}. Note that our pipeline does not predict frame-wise motion recursively. Instead, the batch prediction of flow maps alleviates the effect of accumulated error and possible blur artifacts.

\vspace{2mm}
\noindent\textbf{Generative model.}
We propose a conditional generative adversarial network to predict the future optical flow. The pipeline has one generator $\mathcal{G}_{back}$ and two types of discriminators, one for evaluating single frame $\mathcal{D}_f$ and the other for temporal coherence of multiple video frames $\mathcal{D}_v$. 
The generator $\mathcal{G}_{back}$ is an encoder-decoder structure with skip connections. The encoder follows the structure of ResNet50~\cite{resnet} with activation function replaced with Leaky ReLU~\cite{maas2013rectifier}. 
The input of encoder is a tensor that collates sequential input images $\{x_i\}_{i=1}^{t}$, sequential semantic layout $\{s_i\}_{i=1}^{t}$ generated by~\cite{semantic_cvpr19}, sequential instance maps $\{e_i\}_{i=1}^{t}$ computed by \cite{xiong19upsnet}, and sequential optical flow between consecutive input frames $\{f_i\}_{i=1}^{t-1}$ using PWC-Net~\cite{Sun2018PWC-Net}. 

The decoder consists of several upsample modules. We employ the multi-scale strategy to predict optical flow at different spatial resolutions. The input to each module is a concatenation of feature maps produced at the corresponding resolution by the encoder, feature maps provided by the preceding module, and optical flow prediction result. Each upsample module consist of a bilinear upsample layer and a convolutional layer to recover the spatial resolution.

A loss function of frame discriminator $\mathcal{L}_f$ checks if estimated flow creates weird artifacts by warping $x_t$ using predicted optical flow:
\begin{equation}
    \mathcal{L}_f =\sum_{i=t+1}^{T}\Big(\mathrm{log} \mathcal{D}_f(x_i, f_{i\rightarrow t}) + \mathrm{log}(1 - \mathcal{D}_f\big(\Tilde{x}_{i}, \Tilde{f}_{i\rightarrow t})\big)\Big),
\end{equation}
where $\Tilde{f}_{i\rightarrow t}$ is the optical flow from frame $i$ to frame $t$ predicted by $\mathcal{G}_{back}$, $\Tilde{x}_{i}$ is inversely warped image of $x_t$ using $\Tilde{f}_{i\rightarrow t}$, and $f_{i\rightarrow t}$ is the ground-truth optical flow from frame $i$ to frame $t$. The loss $\mathcal{L}_v$ on $\mathcal{D}_v$ is defined as:
\begin{align}
    \mathcal{L}_v~=~& \mathrm{log}\mathcal{D}_v\big(\{x_i\}_{i=1}^{T}, \{f_{i\rightarrow t}\}_{i=t+1}^{T}\big)~+~\\ \nonumber
    &\mathrm{log}\Big(1- \mathcal{D}_v\big(\{\Tilde{x_i}\}_{i=1}^{T}, \{\Tilde{f}_{i\rightarrow t}\}_{i=t+1}^{T}\big)\Big),
\end{align}
where $\{x_i\}_{i=1}^{T}$ concatenates images $\{x_{1},\cdots,x_{T}\}$ in the channel-wise manner, $\{f_{i\rightarrow t}\}_{i=t+1}^{T}$ is concatenated optical flow, and others are defined similarly.
In contrast to $\mathcal{L}_f$, this function penalizes unrealistic image and motion by directly analyzing a range of image frames and flow maps. This is realized by concatenating frames to learn temporal changes. In this way, unrealistic temporal behavior is discouraged.

\vspace{2mm}
\noindent\textbf{Flow evaluation.}
We have an additive loss $\mathcal{L}_{flow}$ to evaluate estimated flow.
$\mathcal{L}_{flow}$ is linear combination of multiple criterions $\mathcal{L}_{flow} = \sum (\lambda_{data}\mathcal{L}_{data} + \lambda_{perc}\mathcal{L}_{perc} + \lambda_{smooth}\mathcal{L}_{smooth} + \lambda_{cons}\mathcal{L}_{cons})$, where ($\lambda_{data}, \lambda_{perc}, \lambda_{smooth}, \lambda_{cons}$) is empirically set to (1.0, 15.0, 1.0, 1.0), respectively.

$\mathcal{L}_{data}$ is a data term that penalizes the discrepancy between predicted flow and the flow from real images:
\begin{equation}
\mathcal{L}_{data} = \sum_{i=t+1}^{T}{C_{i\shortrightarrow{t}}\left \| \Tilde{f}_{i\shortrightarrow{t}}  - f_{i\shortrightarrow{t}}  \right \|_1},
\end{equation}
where a confidence map $C$ indicates whether the optical flow on this pixel is valid. 

We also compute a perceptual loss between warped image and ground truth image. We use VGG19 model \cite{vgg} for feature extraction and define a $L_1$ loss between warped images and ground truth images in the feature domain:
\begin{equation}
\mathcal{L}_{perc} = \sum_{i=t+1}^{T}\left ( \sum_{j=1}^{n}\frac{1}{N_j}\left \| \Phi_j(\tilde{x}_i) -\Phi_j(x_i) \right \|_1 \right ),
\end{equation}
where $n$ is the number of VGG feature layers. where $\Phi_j$ denote feature map from the $j$-th layer in the VGG-19 network having a number of feature parameter $N_j$.
To make the predicted optical flow coherent with the structure of $x_i$, we adopt smoothness loss for optical flow weighted by image gradient $\nabla x_i$:
\begin{equation}
    \mathcal{L}_{smooth} = \sum_{i=t+1}^{T}\left\| \nabla \tilde{f}_{i\rightarrow t} \right\|_1 e^{-\left\| \nabla x_i \right \|_1},
\end{equation}
where $\nabla$ indicates the gradient operator. 

To make the training more stable, we use a forward-backward consistency loss~\cite{yin2018geonet}:
\begin{equation}
    \mathcal{L}_{cons} = \sum_{i=t+1}^{T} \sum_\mathbf{p} \delta(\mathbf{p})  \left \| \Delta \Tilde{f}_{i \rightarrow t}(\mathbf{p}) \right \|_1,
\end{equation}
where $\Delta \Tilde{f}_{i \rightarrow t}(\mathbf{p})$ is the discrepancy obtained from forward and backward flow check at pixel location $\mathbf{p}$. It is defined as $\Delta \Tilde{f}_{i \rightarrow t}(\mathbf{p}) = \mathbf{p}- \big(\mathbf{p}' + \Tilde{f}_{t \rightarrow i}(\mathbf{p}')\big)$, where $\mathbf{p}'=\mathbf{p} + \Tilde{f}_{i \rightarrow t}(\mathbf{p})$. 
$\delta(\mathbf{p})$ is a conditional scalar for robustness. $\delta(\mathbf{p})$ is 1 if $\left \| \Delta \Tilde{f}_{i\rightarrow t}(\mathbf{p}) \right \|_2 < \max\big( a, b \left \| \Tilde{f}_{i\rightarrow t}(\mathbf{p}) \right \|_2 \big)$ or 0 otherwise. $(a, b)$ is empirically set to (3, 0.05). 
Pixels where the forward and backward flows contradict seriously are regarded as possible outliers.

As a result, we train the flow prediction network using a combination of proposed losses\footnote{We also define the similar losses with opposite flow direction to improve consistency.}:
\begin{equation}
    \underset{\mathcal{G}_{back}}{\mathrm{min}}\big(\underset{\mathcal{D}_f}{\mathrm{max}}\lambda_f\mathcal{L}_f + \underset{\mathcal{D}_v}{\mathrm{max}}\lambda_v\mathcal{L}_v + \mathcal{L}_{flow}\big).
\end{equation}
The weight for frame discriminator $\lambda_f$ and video discriminator $\lambda_v$ is empirically set to 1.0 and 2.0. Here we use the multi-scale loss that is defined as the sum of the losses when images are evaluated at different resolutions: full resolution, half resolution,  $\frac{1}{4}$ resolution, and so on.

\vspace{2mm}
\noindent\textbf{Background inpainting.}
For better future prediction, we decompose moving objects and static scenes from the input image. After extracting moving objects, the area where moving objects were placed remains blank. Such a blank region is filled with an inpainting network based on Wasserstein GANs with a contextual attention layer~\cite{yu2018generative}. To make the inpainting network even better, we feed randomly cropped patches from background classes (such as buildings, trees, or roads in traffic scenes) and perform fine-tuning. This procedure makes an inpainted background image $b_i$ from the original image with holes.

The background inpainting operation is necessary. It is because the inpainted background is used as the extra guidance for dynamic object trajectories prediction. Without background inpainting, the regions for moving objects are denoted as black. Then the dynamic object trajectories prediction module will overfit to predict motion to match the black pixels, which is not desirable.

\begin{figure}[t!]
\centering
\hspace{-2mm}
\includegraphics[width=1\linewidth]{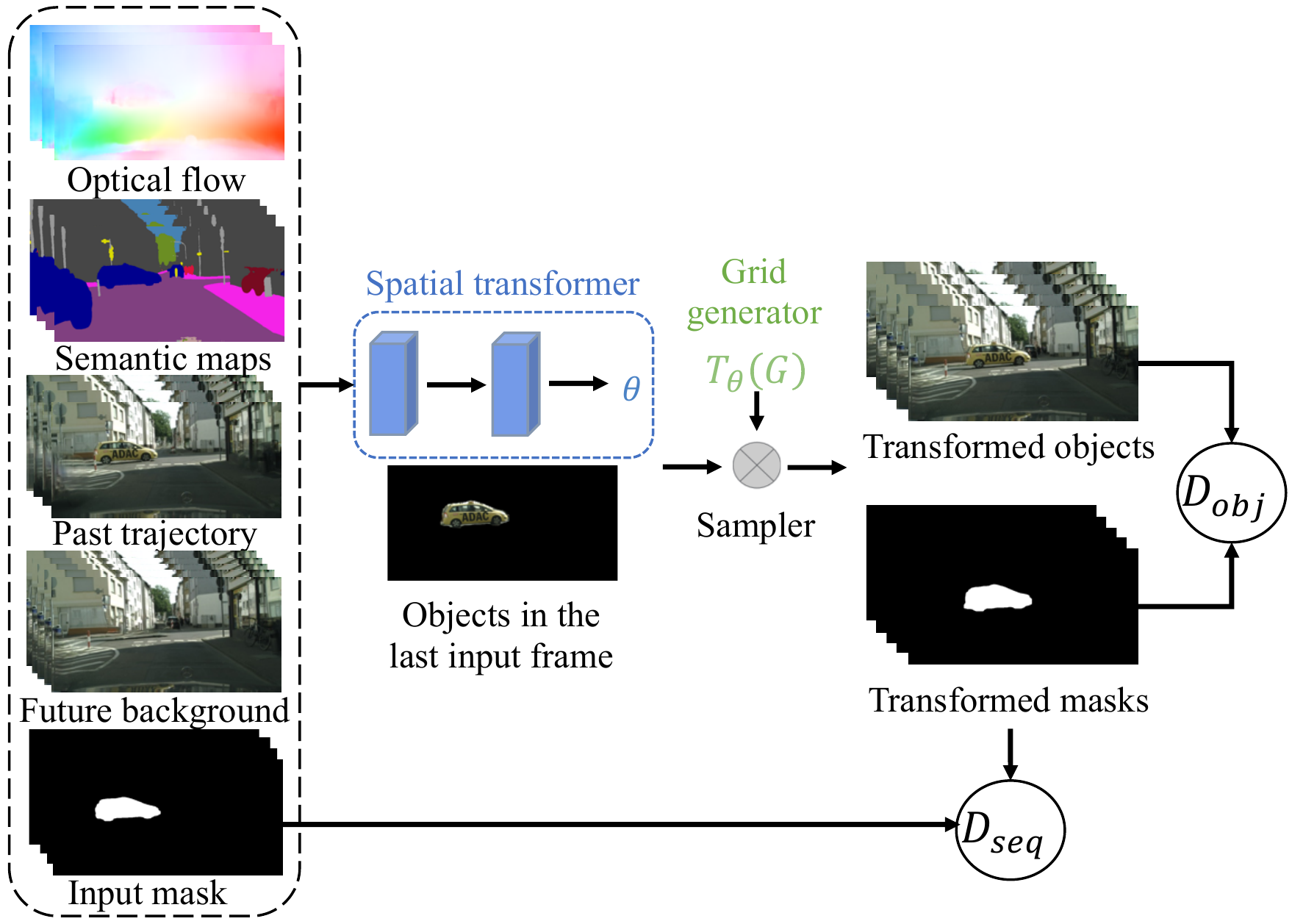}
\vspace{1mm}
\caption{
Training loss for dynamic motion prediction. Our approach puts the predicted objects (by spatial transformer) on the predicted background images and generates virtual images. We also use two discriminators to ensure the locations of predicted objects are spatially and temporally coherent.
}
\label{fig:fore}
\end{figure}

\subsection{Dynamic object motion prediction}
Our approach identifies dynamic objects in the scene and handles their motion explicitly. Instead of treating cluttered scenes as a whole, this scheme helps to understand the history of an individual object so that it can predict the future better. We presume the motion of dynamic objects can be adequately approximated with 2D affine transformation. Due to this rigid motion constraint, predicted appearance does not show distortion or unrealistic texture that are common problems in previous approaches. Our model detects all the moving objects, and each object is treated separately using our transformation network.

\vspace{2mm}
\noindent\textbf{Network.}
The input to the motion prediction network is a sequence of binary object masks $m$, optical flow $f$, semantic maps $s$, objects $o$, and inpainted background images $b$. The network produces a series of 2D affine transformation $\mathbf{A}$ that expresses the predicted object motion. Note that the network takes background images as input because the location of objects is highly related to the background. For example, a car should be placed on the road, and trees should not block the road, etc. Without background information, the network may predict unrealistic trajectories because the prediction is purely based on past motions.

The network is an encoder architecture and outputs the parameters of a series of 2D affine transformation $\mathbf{A}$. 
Then the following grid sampler transforms coordinate of object's pixels in the last frame using the estimated parameters. By combining the estimated background image $b$ and transformed object $o$, we can build a composition image $c$.

Similar to the background prediction module, the motion prediction network is equipped with two discriminators: single object discriminator $\mathcal{D}_{obj}$ and object sequence discriminator $\mathcal{D}_{seq}$.
The input of $\mathcal{D}_{obj}$ is a pair of an object mask and a composed image to determine whether the predicted location is natural. This discriminator is used to suppress some unreasonable areas, such as cars on a building. The produced image is made by placing a transformed object on an inpainted background. The input of $\mathcal{D}_{seq}$ takes a sequence of masks representing the object trajectory as input and determines whether the predicted object trajectory is reasonable.

We define the discriminator loss $\mathcal{L}_{obj}$ on single object discriminator and the discriminator loss $\mathcal{L}_{seq}$ on object sequence discriminator as follows:
\begin{equation}
    \mathcal{L}_{obj}=\sum_{i=t+1}^{T}\Big(\mathrm{log} \mathcal{D}_{obj}(c_i, m_i) + \mathrm{log}\big(1 - \mathcal{D}_{obj}(\Tilde{c_i}, \Tilde{m_i})\big)\Big),
\end{equation}
\begin{align}
    \mathcal{L}_{seq}=~&~\mathrm{log}\mathcal{D}_{seq}\big(\{m_i\}_{i=1}^{T}\big)+\mathrm{log}\Big(1-\mathcal{D}_{seq}\big(\{\Tilde{m_i}\}_{i=1}^{T}\big)\Big),
\end{align}
where $\mathcal{L}_{obj}$ is the GAN loss on mask and synthetic image pair defined by single object discriminator $\mathcal{D}_{obj}$, $\mathcal{L}_{seq}$ is the GAN loss on sequential masks defined by object sequence discriminator $\mathcal{D}_{seq}$. $\Tilde{c}_i$ is the composite of transformed object and background information, and $\Tilde{m}_i$ is a binary mask of moving object.

Another loss $\mathcal{L}_r$ consists of three terms, and it is equivalent to $\lambda_{rgb} \mathcal{L}_{rgb} + \lambda_{reg} \mathcal{L}_{reg} + \lambda_{smooth} \mathcal{L}_{smooth}$, where $(\lambda_{rgb}, \lambda_{reg}, \lambda_{smooth})$ is set to (1.0, 1.0, 2.0).
$\mathcal{L}_{rgb}$ is the $L_1$ difference between appearance of a $j$-th object in $i$-th frame and its ground truth: 
$\mathcal{L}_{rgb} = \sum_{i=t+1}^{T} m_{(i,j)} \odot \left \|\Tilde{o}_{(i,j)} - o_{(i,j)} \right \|_1$, where $\Tilde{o}$ is transformed object. 
$\mathcal{L}_{smooth}$ is the smoothness loss to improve the temporal coherency of predicted parameters: $\mathcal{L}_{smooth} = \sum_{i=t+3}^{T}\sum_{j} \|\big(\mathbf{A}_{(i,j)}-\mathbf{A}_{({i-1},j)}\big) - \big(\mathbf{A}_{({i-1},j)}-\mathbf{A}_{({i-2},j)}\big) \|_1$.

$\mathcal{L}_{reg}$ is a regularization term on predicted parameters to prevent abrupt change from original state, or identity transform $\mathbf{I}$:
$\mathcal{L}_{reg} = \sum_{i=t+1}^{T}\sum_{j} \|\mathbf{A}_{(i,j)}-\mathbf{I}\|_2$.

As a result, for each moving object, we a separate motion estimation network and the loss for training this network is:
\begin{equation}
    \underset{\mathcal{G}_{fore}}{\mathrm{min}}(\underset{\mathcal{D}_{obj}}{\mathrm{max}}\lambda_{obj}\mathcal{L}_{obj} + \underset{\mathcal{D}_{seq}}{\mathrm{max}}\lambda_{seq}\mathcal{L}_{seq} + \lambda_{r}\mathcal{L}_r),
\end{equation}
where $(\lambda_{obj}, \lambda_{seq}, \lambda_{r})$ is set to $(4.0, 4.0, 1.0)$ and $\mathcal{G}_{fore}$ is the foreground object generator.

\vspace{2mm}
\noindent\textbf{Training data generation.}
The Cityscapes dataset~\cite{Cordts2016Cityscapes} and KITTI dataset~\cite{kitti} does not provide tracking information for each instance. Therefore, we employ a tracking algorithm to produce data for training the proposed network. We first generate an instance mask using the approach by Xiong~\etal~\cite{xiong19upsnet}. Then, few-shot tracking algorithm~\cite{li2018siamrpn++} is employed to obtain bounding boxes of the tracked objects in a video sequence. After getting the bounding boxes of the tracked objects, we compute the intersection of bounding boxes and instance maps to obtain the corresponding binary masks. We employ several strategies to delete some failure tracking samples. For instance, we compute the SSIM~\cite{ssim} score of objects being tracked to determine whether they are the same object.

\subsection{Background-foreground composition}
After predicting motion for the background scene and moving objects, the composition module fuses the scene components to create future video frames. We determine the relative depth order of moving objects according to the relative depth obtained by GeoNet~\cite{yin2018geonet}. Then we place the moving objects one by one onto the predicted background. Note that we have a hole-filled background image $b_i$, we may directly use those frames for producing output, but it does not have temporal coherence. 

Therefore, we adopt a video inpainting approach to minimize flickering artifact. Following the method~\cite{Xu_2019_CVPR}, we utilize forward and backward optical flow between consecutive frames, and employ a consistency check to find valid optical flow. With adequate optical flow, we build a connection between pixels across continuous frames. Pixels with the valid flow are propagated bidirectionally to fill the missing regions. 
This procedure repeats to minimize holes in the video. If there are still missing regions, the image inpainting method~\cite{yu2018generative} is employed to fill such areas.

\begin{figure*}[t!]
\hspace{-4mm}
\setlength\tabcolsep{2pt}
\begin{tabular}{cccc}
\rotatebox{90}{\hspace{6mm}$t+1$} & \includegraphics[width=0.33\linewidth]{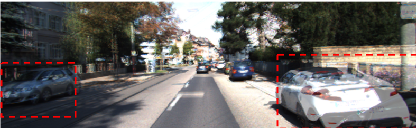}  & \includegraphics[width=0.33\linewidth]{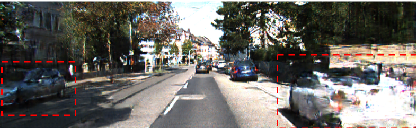}  & \includegraphics[width=0.33\linewidth]{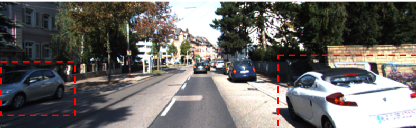} \\
\rotatebox{90}{\hspace{6mm}$t+3$ } & \includegraphics[width=0.33\linewidth]{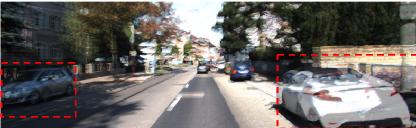}  & \includegraphics[width=0.33\linewidth]{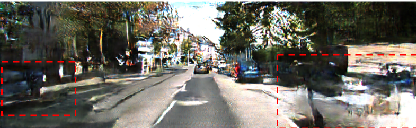}  & 
\includegraphics[width=0.33\linewidth]{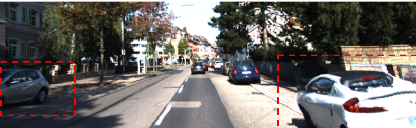} \\
\rotatebox{90}{\hspace{6mm}$t+5$} & \includegraphics[width=0.33\linewidth]{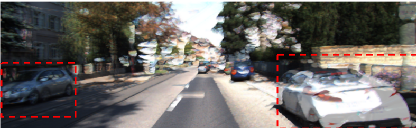}   & \includegraphics[width=0.33\linewidth]{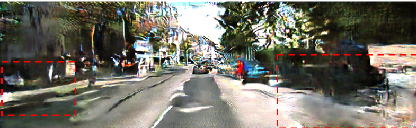}   & 
\includegraphics[width=0.33\linewidth]{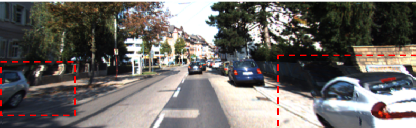} \\
\rotatebox{90}{\hspace{6mm}$t+1$} & \includegraphics[width=0.33\linewidth]{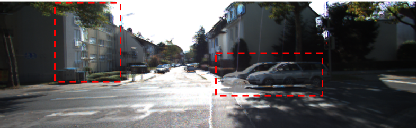}  & \includegraphics[width=0.33\linewidth]{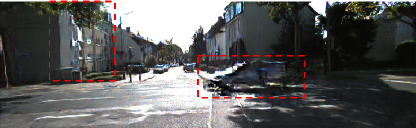}  & \includegraphics[width=0.33\linewidth]{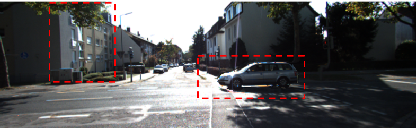} \\
\rotatebox{90}{\hspace{6mm}$t+3$ } & \includegraphics[width=0.33\linewidth]{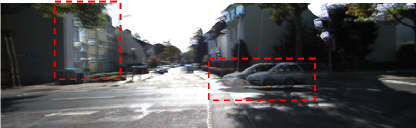}  & \includegraphics[width=0.33\linewidth]{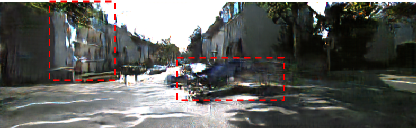}  & 
\includegraphics[width=0.33\linewidth]{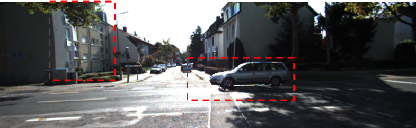} \\
\rotatebox{90}{\hspace{6mm}$t+5$} & \includegraphics[width=0.33\linewidth]{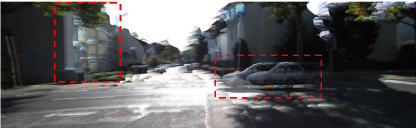} & \includegraphics[width=0.33\linewidth]{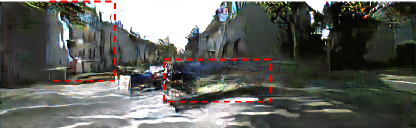}   & 
\includegraphics[width=0.33\linewidth]{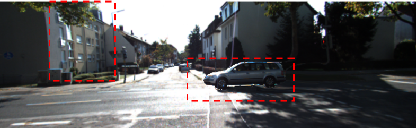} \\
& Voxel-Flow~\cite{liu2017voxelflow} & MCNet~\cite{villegas17mcnet} & Ours \\
\end{tabular}
\caption{Results of predicting the frames $t+1$, $t+3$ , and $t+5$ on the KITTI dataset~\cite{kitti}.}\label{fig:kitti_compare}
\end{figure*}

\section{Experiments}

\begin{table*}
\centering
\resizebox{0.9995\linewidth}{!}{
\ra{1.25}
\begin{tabular}{ccccccccccccc}
\toprule
& \multicolumn{6}{c}{Cityscapes} & \multicolumn{6}{c}{KITTI} \\
\midrule
& \multicolumn{2}{c}{Next frame} & \multicolumn{2}{c}{Next 5 frames} & \multicolumn{2}{c}{Next 10 frames} & \multicolumn{2}{c}{Next frame} & \multicolumn{2}{c}{Next 3 frames} & \multicolumn{2}{c}{Next 5 frames} \\
& MS-SSIM  & LPIPS & MS-SSIM  & LPIPS & MS-SSIM  & LPIPS & MS-SSIM  & LPIPS & MS-SSIM  & LPIPS & MS-SSIM  & LPIPS \\ 
\midrule
PredNet~\cite{prednet} & 0.8403 & 0.2599           & 0.7521 & 0.3603 & 0.6633 & 0.5221  & 0.5626  & 0.5535  & 0.5147 & 0.5866  & 0.4756 & 0.6295                \\
MCNET~\cite{villegas17mcnet}       & \textbf{0.8969}  & 0.1888 & 0.7058       & 0.3734  & 0.5971  & 0.4513 & 0.7535  & 0.2405 & 0.6352 & 0.3171 & 0.5548  &  0.3739 \\
Voxel Flow~\cite{liu2017voxelflow} & 0.8385 & 0.1737 & 0.7111 & 0.2879 & 0.6341 & 0.3655  & 0.5393   & 0.3247  & 0.4699  & 0.3743 & 0.4262  & 0.4159        \\
Vid2vid~\cite{Wang2018}            & 0.8816      & 0.1058 & 0.7513 & 0.2014 & 0.6690  & 0.2705  & - & - & - & - & - & - \\
\midrule
Ours-WC            & 0.8792 & 0.0903  & 0.7430  & 0.1718  & 0.6593 & 0.2411 & 0.6853  & 0.2252  & 0.5850  & 0.2897  & 0.5217 & 0.3482 \\
Ours-WM            & 0.8866 & 0.0899  & 0.7537  & 0.1694  & 0.6727 & 0.2351 & 0.7634  & 0.1987  & 0.6504  & 0.2588  & 0.5839 & 0.3136      \\
\hline
Ours          & 0.8910  & \textbf{0.0850} & \textbf{0.7568} & \textbf{0.1650} & \textbf{0.6741} & \textbf{0.2328}  & \textbf{0.7928} &  \textbf{0.1848}  & \textbf{0.6765}  & \textbf{0.2461}   & \textbf{0.6077}  &  \textbf{0.3049}\\
\bottomrule
\end{tabular}
}
\vspace{2mm}
\caption{Comparison with state-of-the-art methods on the Cityscapes and KITTI datasets. The table shows the image quality of the synthesized images. The higher MS-SSIM is better. The lower LPIPS is better.}
\label{table:whole_compare}
\end{table*}

We conduct both quantitative and qualitative experiments on real-world datasets concerning the capability of predicting future video. We compare our approach with other approaches that produce the next-frame or multiple-frames for the future.

\subsection{Datasets}
We conducted our experiments on Cityscapes dataset~\cite{Cordts2016Cityscapes} and KITTI dataset ~\cite{kitti}. Cityscapes dataset contains $2048\times1024$ resolution image sequences for city scene captured at 17 FPS. For the fair comparison with other approaches that do not produce such resolution, we experiment at the  $1024\times512$ resolution.
KITTI dataset contains $375\times1242$ resolution image sequences for driving scenes captured at 10 FPS. The semantic maps are generated using the method of ~\cite{semantic_cvpr19}. For the experiment, we get instance maps using UPSNet~\cite{xiong19upsnet} and obtain optical flow fields with PWCNet~\cite{Sun2018PWC-Net}. For the fair comparison, we experiment at the $256\times832$ resolution. We apply techniques such as random horizontal flipping to augment data.

Cityscapes dataset contains 2975 video sequences for training and 500 video sequences for testing. KITTI dataset for our training and evaluation includes 28 video sequences. We randomly select four sequences for assessment. 

\subsection{Implementation}
We use the multi-scale PatchGAN discriminator~\cite{pix2pix2017} architecture for all the discriminators in our framework.
For the Cityscapes dataset, the input frame length is set to 4, and the prediction length is set to 5. We first train a model at the $256\times512$ resolution, then train a $512\times1024$ resolution model by adding an upsampling module. By recurrently test our model twice, we obtain future predictions for the next 10 frames.

For the KITTI dataset, the input frame length is set to 4. Because the KITTI dataset has a more substantial motion, generating optical flow between two long period frame is difficult using PWCNet~\cite{Sun2018PWC-Net}. The prediction length for the background prediction model is set to 3. And the prediction length for the dynamic object motion prediction model is set to 5. We experiment at the $256\times832$ resolution. By recurrently test the model twice, we obtain predicted images in the next 5 frames.

All parts of our model are implemented with Pytorch 1.1.0, and we use the ADAM optimizer. For the background prediction model, we train 200 epochs, with learning rate 2e-4 for the first 100 epochs, then linearly decrease the learning rate. For the dynamic trajectory prediction model, we train 60 epochs with a learning rate of 3e-5. Training takes about three days for a $512\times1024$ resolution model. The experiment is done with Nvidia RTX 2080 Ti. 
 
\subsection{Evaluation metrics}
We evaluate our model using several metrics measuring the accuracy of video frames in the future. We use a multi-scale structure similarity (MS-SSIM)~\cite{msssim} index and perceptual image patch similarity (LPIPS)~\cite{lpips}. Higher MS-SSIM scores and lower LPIPS distances suggest better performance.

\subsection{Baselines}
To evaluate our model for future prediction, we compare our model with the following baselines, where the first several are state-of-the-art approaches, and the rest are variants of our model.

\vspace{2mm}
\noindent \textbf{PredNet}~\cite{prednet}. PredNet is a prior approach for next-frame prediction. We fine-tune their model on our dataset, and recurrently perform next-frame prediction to get multiple-frame results.

\vspace{2mm}
\noindent \textbf{MCNet}~\cite{villegas17mcnet}. This is a state-of-the-art approach for the next frame prediction. We re-train their model on our datasets using their public source code. The multiple frames are generated by recurrently applying the pipeline.

\vspace{2mm}
\noindent \textbf{Voxel-Flow}~\cite{liu2017voxelflow}. This is a video synthesis approach with optical flow fields across space and time. This approach can be applied for video extrapolation. We re-train their model on our dataset for evaluation.

\vspace{2mm}
\noindent \textbf{Vid2vid}~\cite{Wang2018}. This is a video-to-video translation framework. The approach can generate a video conditioned on a sequence of semantic layouts. For future prediction, their approach predicts the semantic layout and converts a sequence of semantic layouts into a real video. We directly compare our method with their provided video prediction results on the Cityscapes dataset.

\vspace{2mm}
\noindent \textbf{Ours-WC.} Our ablated model without foreground-background composition. To demonstrate the effectiveness of our foreground-background separation approach, we train a model to directly output the optical flow prediction for a full image using the same model with the background prediction.

\vspace{2mm}
\noindent \textbf{Ours-WM.} Our ablated model without moving object detection. For this model, we remove the moving object detection module and use an STN to predict the trajectories of all possible moving objects (cars, pedestrians) based on semantic classes.

\subsection{Evaluation on Cityscapes and KITTI}
We evaluate the capability of our model to predict future video frames in both the next-frame and multiple-frames prediction. Our result on Cityscapes and KITTI dataset is shown in Table~\ref{table:whole_compare}. The frame rates of Cityscapes dataset and KITTI dataset are 17 FPS and 10 FPS, respectively. Then we predict the next 10 frames on the Cityscapes dataset and the next 5 frames on the KITTI dataset, about 0.5 seconds.

On the Cityscapes dataset, in terms of MS-SSIM score, our model achieves comparable scores with MCNet~\cite{villegas17mcnet} and vid2vid~\cite{Wang2018}. The performance of our model in LPIPS is 20\%, 18\%, 14\% better than the second-best model for the evaluation of the next frame, next five frames, and the next ten frames. On the KITTI dataset, our model outperforms all state-of-the-art methods in all metrics. Our model's improvement in terms of LPIPS in the next frame, next three frames, next five frames, is 23\%, 22\%, 18\%, respectively against the second-best result. Our improvement in terms of MS-SSIM is 5\%, 7\%, 10\%, respectively.
It demonstrates that our method can achieve better performance in both short-term and long term prediction. It is because our approach highly keeps the rigidity of objects. The current state-of-the-art method appears to have significant distortion artifact around object boundary, while our approach alleviates this phenomenon a lot and makes the result more realistic.

We also perform an ablation study on Ours-WC and Ours-WM. From the results, we can see that all the strategies in our model are helpful. The foreground-background decomposition keeps the rigidity of objects and makes the background prediction easier. The moving object detection strategy classifies objects into dynamic or static and predicts separately based on the motion type.

As demonstrated in Fig.~\ref{fig:cityscapes_compare} and ~\ref{fig:kitti_compare}, our model produces more realistic results over state-of-the-art methods. Our method keeps the rigidity of objects even in long-term prediction, while the state-of-the-art techniques suffer from distortion around motion boundaries. Also, our method produces a result with less blurriness because we predict the motion of multiple frames together. This strategy alleviates the accumulated error by recurrent prediction. More visual comparisons are shown in the supplement.

\subsection{Additional experiments}
We also conduct experiments beyond driving scenes on the BAIR robot pushing dataset~\cite{Finn2016} and the Penn Action dataset~\cite{penn_action}. The BAIR dataset consists of videos about a robot arm pushing multiple objects. The Penn dataset has videos with various non-rigid human actions. The results are presented in the supplement.

\section{Conclusion}
We have presented a separate-predict-composite model for future frame prediction. Our method produces future frames by firstly decomposing possible moving objects into currently-moving or static objects. Then for moving objects, we employ a spatial transformer network to predict the trajectories of objects. This helps to preserve the structure of objects while producing reliable future motion. For background, we use an optical flow prediction network to predict the background of multiple frames at once. Then we integrate the foreground and background and add a video in-painting module to help alleviate the artifact in composition. The experiments have shown that our approach outperforms prior work on future video prediction.

\clearpage
\newpage
\balance
{\small
\bibliographystyle{ieee_fullname}
\bibliography{reference}
}

\end{document}